\documentclass[conference]{IEEEtran}
\IEEEoverridecommandlockouts
\usepackage{cite}
\usepackage{amsmath,amssymb,amsfonts}
\usepackage{algorithmic}
\usepackage{graphicx}
\usepackage{textcomp}
\usepackage{xcolor}
\usepackage{booktabs}
\usepackage{multirow}

\def\BibTeX{{\rm B\kern-.05em{\sc i\kern-.025em b}\kern-.08em
    T\kern-.1667em\lower.7ex\hbox{E}\kern-.125emX}}
\begin{document}

\title{HyperPriv-EPN: Hypergraph Learning with Privileged Knowledge for Ependymoma Prognosis\\
}

\author{\IEEEauthorblockN{1\textsuperscript{st} Shuren Gabriel Yu}
\IEEEauthorblockA{\textit{School of Software, Tsinghua University} \\
Beijing, China \\
yusr22@mails.tsinghua.edu.cn}
\and

\IEEEauthorblockN{2\textsuperscript{nd} Sikang Ren}
\IEEEauthorblockA{\textit{Beijing Tiantan Hospital} \\
Beijing, China \\
rensk777@163.com}
\and

\IEEEauthorblockN{3\textsuperscript{nd} Yongji Tian}
\IEEEauthorblockA{\textit{Beijing Tiantan Hospital} \\
Beijing, China \\
tianyongji@bjtth.org}
}

\maketitle

\begin{abstract}
Preoperative prognosis of Ependymoma is critical for treatment planning but challenging due to the lack of semantic insights in MRI compared to post-operative surgical reports. Existing multimodal methods fail to leverage this privileged text data when it is unavailable during inference. To bridge this gap, we propose HyperPriv-EPN, a hypergraph-based Learning Using Privileged Information (LUPI) framework. We introduce a Severed Graph Strategy, utilizing a shared encoder to process both a Teacher graph (enriched with privileged post-surgery information) and a Student graph (restricted to pre-operation data). Through dual-stream distillation, the Student learns to hallucinate semantic community structures from visual features alone. Validated on a multi-center cohort of 311 patients, HyperPriv-EPN achieves state-of-the-art diagnostic accuracy and survival stratification. This effectively transfers expert knowledge to the preoperative setting, unlocking the value of historical post-operative data to guide the diagnosis of new patients without requiring text at inference.
\end{abstract}


\section{Introduction}
Ependymoma is a heterogeneous central nervous system (CNS) malignancy that exhibits significant variability in molecular profiles and clinical outcomes. Under the 2021 World Health Organization (WHO) CNS classification \cite{louis2021who}, accurate preoperative assessment—specifically predicting molecular groups (e.g., Posterior Fossa A vs. B)\cite{pajtler2015molecular} and WHO grades—is critical for risk stratification and therapeutic decision-making. While histopathology remains the gold standard, it is invasive and carries surgical risks. Consequently, multi-sequence Magnetic Resonance Imaging (MRI) has emerged as the primary non-invasive modality for preoperative diagnosis.

Despite the success of deep learning in medical imaging, existing MRI-based prognostic systems face two fundamental limitations. First, the "Independence Assumption" ignores population-level wisdom. Standard Convolutional Neural Networks (CNNs) and attention-based Multiple Instance Learning (MIL) approaches, such as AttentionDeepMIL\cite{ilse2018attention}, process each patient as an isolated instance. This independent modeling fails to capture high-order correlations, such as the latent structural similarities shared by a "community" of patients with specific aggressive molecular subtypes.

Second, there is a critical "Inference-Time Modality Gap." In retrospective training cohorts, rich multimodal data is often available, including detailed radiology reports and surgical notes that explicitly describe biological attributes (e.g., "necrosis" or "vascular proliferation"). However, existing models typically discard this textual data because it is unavailable for new patients at the time of initial inference (i.e., before surgery). This results in a massive waste of "Privileged Information" (PI)—semantic expert knowledge that could guide the interpretation of ambiguous MRI features. We contend that this data loss is architectural rather than fundamental; the semantic richness of these "future" reports can be effectively engineered into the pre-operative model via distillation. By treating the privileged text as a training scaffold rather than a required input, we can embed these expert insights directly into the visual backbone, making them accessible even when the text itself is absent. While recent multimodal transformers like MCAT\cite{chen2021multimodal} attempt to fuse these modalities, they rely on the presence of both inputs and fail when the text branch is missing during inference.

To bridge this gap, we propose HyperPriv-EPN, a novel Hypergraph Neural Network framework that leverages Learning Using Privileged Information (LUPI). Our key insight is that text is the architect of the hypergraph. While it is difficult to cluster patients solely based on noisy MRI pixels, privileged text reports provide ground-truth semantic concepts that define patient communities. By constructing a Teacher Hypergraph driven by text-defined communities ("Concept Hyperedges"), we provide a robust structural scaffold. We then employ a Severed Graph Distillation strategy to force an MRI-only Student Hypergraph to mimic these high-order topological structures. This effectively allows the Student model to "hallucinate" the missing semantic context from MRI features alone, enabling high-performance inference even when clinical reports are unavailable.

Our main contributions are summarized as follows:

\begin{itemize}
    \item We propose the first hypergraph distillation framework for Ependymoma that explicitly models high-order patient correlations by treating clinical concepts as hyperedges, capturing population-level patterns that pairwise graphs miss.
    \item We introduce a Severed Graph Distillation mechanism that physically removes privileged text connections during the student pass, ensuring that while the model benefits from text during training, the final inference strictly relies on MRI and clinical data to prevent data leakage.
    \item Validated on a multi-center cohort, HyperPriv-EPN significantly outperforms state-of-the-art baselines like MCAT and AttentionDeepMIL in diagnostic tasks (WHO Grade and Molecular Grouping) and Overall Survival (OS) prediction, while achieving competitive performance on par with baselines for Progression-Free Survival (PFS).
\end{itemize}

\section{Related Work}

\subsection{Multimodal Survival Analysis and MIL}
Deep survival analysis has evolved from simple MLP-based approaches like DeepSurv \cite{katzman2018deepsurv} to complex Multiple Instance Learning (MIL) frameworks. Methods such as AttentionDeepMIL \cite{ilse2018attention}, CLAM\cite{lu2021data}, TransMIL\cite{shao2021transmil} and DSMIL \cite{li2021dual} aggregate local patch-level features into patient-level predictions but process each patient independently, ignoring latent population-level correlations. To leverage complementary data, multimodal transformers like MCAT\cite{chen2021multimodal} fuse genomic and histological features using cross-attention mechanisms. However, these fusion frameworks suffer from a rigid "availability assumption": they require all modalities to be present during inference. In clinical practice, where advanced genomic or textual reports are often unavailable pre-operatively, such models fail to deploy effectively.

\subsection{Hypergraph Neural Networks (HGNN)}
Unlike standard Graph Neural Networks (GNNs) that model pairwise relationships — e.g. GCN\cite{kipf2017semi} and GAT \cite{velickovic2018graph} — hypergraphs allow a single edge to connect an arbitrary number of nodes, capturing high-order correlations \cite{feng2019hypergraph}. In medical imaging, HGNNs have been employed to model complex interactions in brain networks and histology patches. However, most existing approaches rely solely on visual similarities to construct hyperedges. Our work extends this by utilizing semantic privileged information to guide the topological construction, ensuring that edges represent biologically meaningful patient communities rather than just visual clusters.

\subsection{Learning Using Privileged Information (LUPI)}
LUPI is a learning paradigm where a teacher model has access to "privileged" data (e.g., text) during training that is unavailable to the student at test time \cite{vapnik2009learning}. This is typically implemented via Knowledge Distillation (KD), where the student is trained to mimic the teacher's representation \cite{hinton2015distilling}. While LUPI has shown promise in action recognition and object detection, its application in medical prognosis is limited. To the best of our knowledge, HyperPriv-EPN is the first framework to apply hypergraph-based LUPI to Ependymoma prognosis, using a "Severed Graph" strategy to distill semantic community structures into a visual-only inference model.

Unlike standard Knowledge Distillation (KD), which typically aims for model compression (distilling a large teacher into a smaller student), our application of LUPI focuses on modality distillation. In this paradigm, the student does not necessarily have fewer parameters than the teacher but operates on a subset of the input space. This distinction is critical in medical imaging, where the "privileged" modality (e.g., histopathology or expert text reports) contains discriminative signals that are fundamentally absent in the student's modality (MRI). Recent works have explored this in uni-modal settings, but our framework extends this to hypergraph topologies, requiring the distillation of not just feature semantics but also structural community relationships.

\section{Method}

\subsection{Problem Formulation and Framework Overview}
We formulate Ependymoma prognosis as a LUPI task. Let $\mathcal{D} = \{(X_i, X^*_i, Y_i)\}_{i=1}^N$ represent a cohort of $N$ patients. Here, $X_i$ denotes the standard modalities available during inference (multi-sequence MRI and clinical attributes), while $X^*_i$ denotes privileged modalities available only during training (surgical pathology reports). $Y_i$ represents the diagnostic labels (Molecular Group, WHO Grade) and survival outcomes (PFS, OS).

\begin{figure*}[ht]
    \centering
    \includegraphics[width=\textwidth]{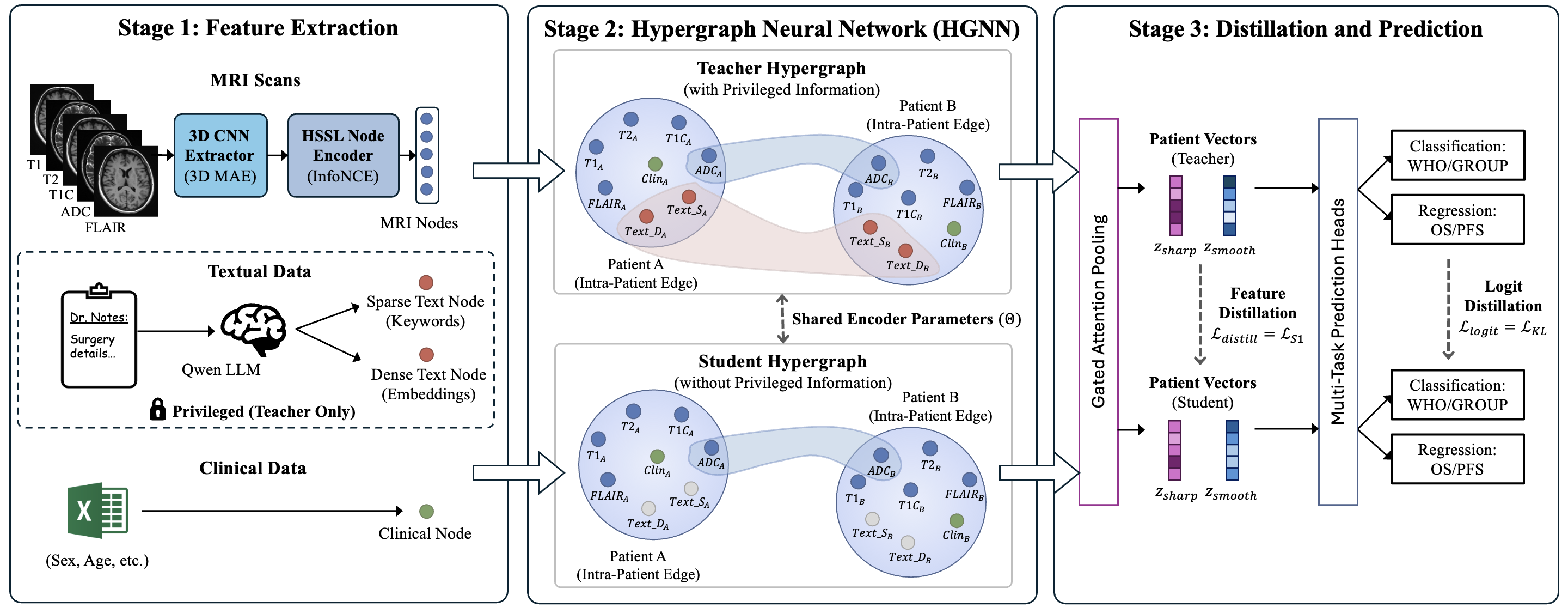}
    \caption{The framework of HyperPriv-EPN. In Stage 1, we extract features from MRI scans, clinical data, and privileged surgery reports to construct multimodal nodes. Stage 2 involves a shared Hypergraph Neural Network encoder that processes both a Teacher hypergraph and a Student hypergraph concurrently. Finally, in Stage 3, node features are aggregated via gated attention pooling into patient-wise vectors, where feature distillation aligns the Student with the Teacher, followed by multi-task prediction heads for classification and regression that utilize logit distillation.}
    \label{fig:framework}
\end{figure*}

As illustrated in Fig. \ref{fig:framework}, our framework, HyperPriv-EPN, aims to learn a student encoder $f_S(X_i)$ that approximates the performance of a teacher encoder $f_T(X_i, X^*_i)$ via knowledge distillation. The framework proceeds in three stages:

\begin{itemize}
    \item \textbf{Stage 1 (Multimodal Extraction):} We encode patient data into a heterogeneous graph with 8 nodes per patient: 5 MRI nodes (processed via 3D-MAE and HSSL), 1 Clinical node, and 2 Privileged Text nodes (processed via Qwen LLM).
    \item \textbf{Stage 2 (Severed Graph Strategy):} We construct two topological variants: a \textit{Teacher Hypergraph} utilizing all nodes to model semantic communities, and a \textit{Student Hypergraph} where text nodes are masked and semantic edges are "severed" to simulate inference.
    \item \textbf{Stage 3 (Distillation):} We employ a shared Hypergraph Neural Network (HGNN) with dual-stream distillation (Feature + Logit) to force the Student to hallucinate the Teacher's semantic topology from visual features alone.
\end{itemize}

\subsection{Hierarchical Feature Extraction}
For each patient $p_{i}$, we extract features into a set of node embeddings $Z_i$.

\subsubsection{Visual Encoding (3D MAE + HSSL)}
To capture lesion-specific heterogeneity from raw MRI sequences (T1, T1C, T2, FLAIR, ADC), we employ a 3D CNN encoder based on a 3D Masked Autoencoder (3D MAE) architecture \cite{he2022masked}. To further refine these features for small-cohort analysis, we utilize a Hierarchical Self-Supervised Learning (HSSL) strategy. A non-linear projection head is trained via an InfoNCE contrastive objective \cite{chen2020simple} to maximize mutual information between volumetric augmentations. The loss function is defined as:

\begin{equation}
\mathcal{L}_{InfoNCE} = -\log \frac{\exp(\text{sim}(z_i, z_i^+) / \tau)}{\sum_{j=1}^{N} \exp(\text{sim}(z_i, z_j) / \tau)}
\end{equation}

This pre-training ensures that the MRI node embeddings $Z_{mri}$ capture discriminative tissue properties (e.g., necrosis vs. edema) prior to hypergraph construction.

\subsubsection{Privileged Text Encoding}
To extract high-quality semantic features from unstructured surgical reports, we employ the Qwen-7B Large Language Model (LLM) \cite{bai2023qwen}, chosen for its strong zero-shot reasoning capabilities in biomedical contexts. We extract two complementary feature representations:
\begin{itemize}
    \item \textbf{Dense Text Node ($h_{dense}$):} We feed the raw surgical text into the LLM and extract the hidden states from the final transformer layer. These dense vectors ($D=4096$) capture the global context and subtle linguistic nuances of the report, such as the degree of tumor resection or descriptions of anatomical invasion.
    \item \textbf{Sparse Concept Node ($h_{sparse}$):} To explicitly model clinical phenotypes, we utilize a prompt-engineering strategy (e.g., "Extract key pathological attributes: necrosis, hemorrhage, calcification..."). The LLM identifies the presence of these specific high-risk keywords, which are encoded into a binary sparse vector. These concepts serve as explicit anchors in the hypergraph, linking patients who share identical biological traits regardless of their visual MRI similarity.
\end{itemize}

\subsection{The Severed Graph Strategy}
To model high-order correlations without data leakage, we do not train separate models. Instead, we employ a single shared HGNN encoder parameterized by $\Theta$ that processes the cohort in two concurrent passes.

\subsubsection{Teacher Pass (Privileged Topology)}
In the first pass, the encoder operates on the full graph $G_T = (\mathcal{V}, \mathcal{E}_{total})$. We define the comprehensive edge set $\mathcal{E}_{total} = \mathcal{E}_{com} \cup \mathcal{E}_{priv}$ as follows:

\textbf{1. Common Edges ($\mathcal{E}_{com}$):} These structures are constructed from non-privileged data and remain active in both Teacher and Student passes.
\begin{itemize}
    \item Intra-Patient Hyperedges: For each patient $p_i$, we construct a hyperedge connecting all their multimodal nodes (MRI, Clinical, and Text). This serves as the primary fusion mechanism for local patient features.
    \item Visual KNN Hyperedges: We compute the cosine similarity between the MRI embeddings of all patients. For each patient $p_i$, we construct hyperedges connecting them to their $k=10$ nearest neighbors in the visual latent space, capturing visually similar tumor morphologies across the cohort.
    \item Clinical KNN Hyperedges: Similarly, we construct hyperedges connecting patients with the $k=10$ most similar clinical feature vectors (e.g., matching age, sex, and tumor location), enforcing demographic and anatomical consistency.
\end{itemize}

\textbf{2. Privileged Edges ($\mathcal{E}_{priv}$):} These structures are derived exclusively from surgical reports and are available only to the Teacher.
\begin{itemize}
    \item Semantic Text KNN: We compute similarities between the dense Qwen embeddings of patient reports. Hyperedges connect patients with semantically related surgical outcomes, capturing subtle linguistic nuances unavailable in the MRI.
    \item Concept Sharing Hyperedges: For each sparse clinical keyword (e.g., "Necrosis"), we form a hyperedge connecting \textit{all} patients whose reports contain that specific term. This explicitly "hard-wires" the ground-truth biological communities into the graph topology.
\end{itemize}

\subsubsection{Student Pass (Severed Topology)}
In the second pass, we simulate the inference scenario by applying a severing operation. We apply a binary mask $M_{blind}$ to strictly zero out all text nodes ($X_{student} = X \odot M_{blind}$), rendering them effectively "dead." Consequently, all edges in $\mathcal{E}_{priv}$ (Text KNN and Concept Sharing) become inactive as their supporting nodes are nulled. The Student is thus forced to propagate information solely through $\mathcal{E}_{com}$ (Intra-patient, Visual KNN, and Clinical KNN). Because the HGNN weights $\Theta$ are shared, the optimizer must learn a visual manifold where patients connected by "Visual KNN" edges also align with the latent semantic communities defined by the now-missing "Concept" edges.

\subsection{Feature Aggregation and Prediction}
To map updated node embeddings into patient-level predictions, we employ a task-specific aggregation strategy. For survival tasks (PFS, OS), we aggregate "smooth" vectors $Z_{smooth}$ using a Gated Attention Mechanism to identify high-risk regions. The importance score $\alpha_k$ for node $h_k$ is computed as:

\begin{equation}
\alpha_k = \frac{\exp\left(\mathbf{w}^T \left(\tanh(\mathbf{V} h_k) \odot \sigma(\mathbf{U} h_k)\right)\right)}{\sum_{j} \exp\left(\mathbf{w}^T \left(\tanh(\mathbf{V} h_j) \odot \sigma(\mathbf{U} h_j)\right)\right)}
\end{equation}

where $\mathbf{w}, \mathbf{V}, \mathbf{U}$ are learnable parameters. The resulting weighted sum $H_{surv}$ effectively "up-weights" critical semantic nodes (e.g., "Necrosis") while suppressing noise. Conversely, for diagnostic classification (Group, WHO), we apply Mean Pooling to the "sharp" visual nodes to obtain $H_{diag}$, ensuring predictions are grounded in global tumor morphology. Finally, these representations are passed to task-specific MLPs: $H_{diag}$ predicts Molecular Group/WHO Grade (optimized via Cross-Entropy), while $H_{surv}$ predicts log-risk scores (optimized via Cox Partial Likelihood).

\subsection{Privileged Knowledge Distillation}
We employ a multi-objective optimization strategy to align the Student's latent representation with the Teacher's "privileged" understanding.

\subsubsection{Hypergraph Propagation \& Disentanglement}
Both graph variants are processed by the shared HGNN. We utilize a Gated Attention Pooling layer to aggregate the active nodes for each patient into a single vector $H_i$. To address the conflicting nature of diagnosis (which requires discriminative, high-frequency features) and prognosis (which requires smooth, ordinal features), we disentangle the output into two vectors: $z_{sharp}$ for classification tasks and $z_{smooth}$ for survival regression.

\subsubsection{Distillation Objectives}
We align the Student and Teacher using a dual-stream loss. First, Feature Distillation forces the Student to "hallucinate" the Teacher's latent position. We minimize the Smooth $L_1$ distance between their respective feature vectors:

\begin{equation}
\begin{split}
\mathcal{L}_{feat} = \frac{1}{N} \sum_{i=1}^{N} [ & \text{SmoothL1}(z_{S,i}^{sharp}, z_{T,i}^{sharp}) \\
& + \text{SmoothL1}(z_{S,i}^{smooth}, z_{T,i}^{smooth}) ]
\end{split}
\end{equation}

Second, logit distillation transfers the Teacher's "soft" confidence (dark knowledge) to the Student. We apply KL-Divergence on the temperature-scaled logits:

\begin{equation}
\mathcal{L}_{logit} = \sum_{k \in \mathcal{T}} \text{KL}\left( \sigma\left(\frac{y_{S,k}}{\tau}\right) \bigg\| \sigma\left(\frac{y_{T,k}}{\tau}\right) \right)
\end{equation}

where $\mathcal{T}$ represents the set of tasks (Group, PFS) and $\tau$ is the temperature parameter.

\subsubsection{Total Objective}
The model is trained end-to-end by minimizing the combined objective:

\begin{equation}
\mathcal{L}_{total} = \mathcal{L}_{task} + \lambda_1 \mathcal{L}_{feat} + \lambda_2 \mathcal{L}_{logit}
\end{equation}

Here, $\mathcal{L}_{task}$ is the summation of task-specific losses: Cross-Entropy for WHO Grading and Molecular Grouping, the Cox Negative Log-Likelihood for survival ranking (PFS/OS), and an auxiliary anatomical regularizer (Cross-Entropy on tumor location) to enforce spatial awareness in the visual encoder.

\section{Experiments}

\begin{figure*}[t]
    \centering
    % --- HEADER ROW: Method Names ---
    % Indent slightly to align with images, skipping the label width
    \hspace{0.03\textwidth} 
    \begin{minipage}{0.18\textwidth} \centering DeepSurv
    \end{minipage}
    \hfill
    \begin{minipage}{0.18\textwidth} \centering Deep-Fusion \end{minipage}
    \hfill
    \begin{minipage}{0.18\textwidth} \centering AttentionDeepMIL \end{minipage}
    \hfill
    \begin{minipage}{0.18\textwidth} \centering MCAT \end{minipage}
    \hfill
    \begin{minipage}{0.18\textwidth} \centering HyperPriv (Ours) \end{minipage}
    
    \vspace{1mm}

    % --- ROW 1: PFS Curves ---
    % Vertical Label (PFS)
    \begin{minipage}{0.03\textwidth}
        \centering
        \rotatebox{90}{PFS}
    \end{minipage}
    \hfill
    % Images
    \begin{minipage}{0.96\textwidth}
        \includegraphics[width=0.19\linewidth]{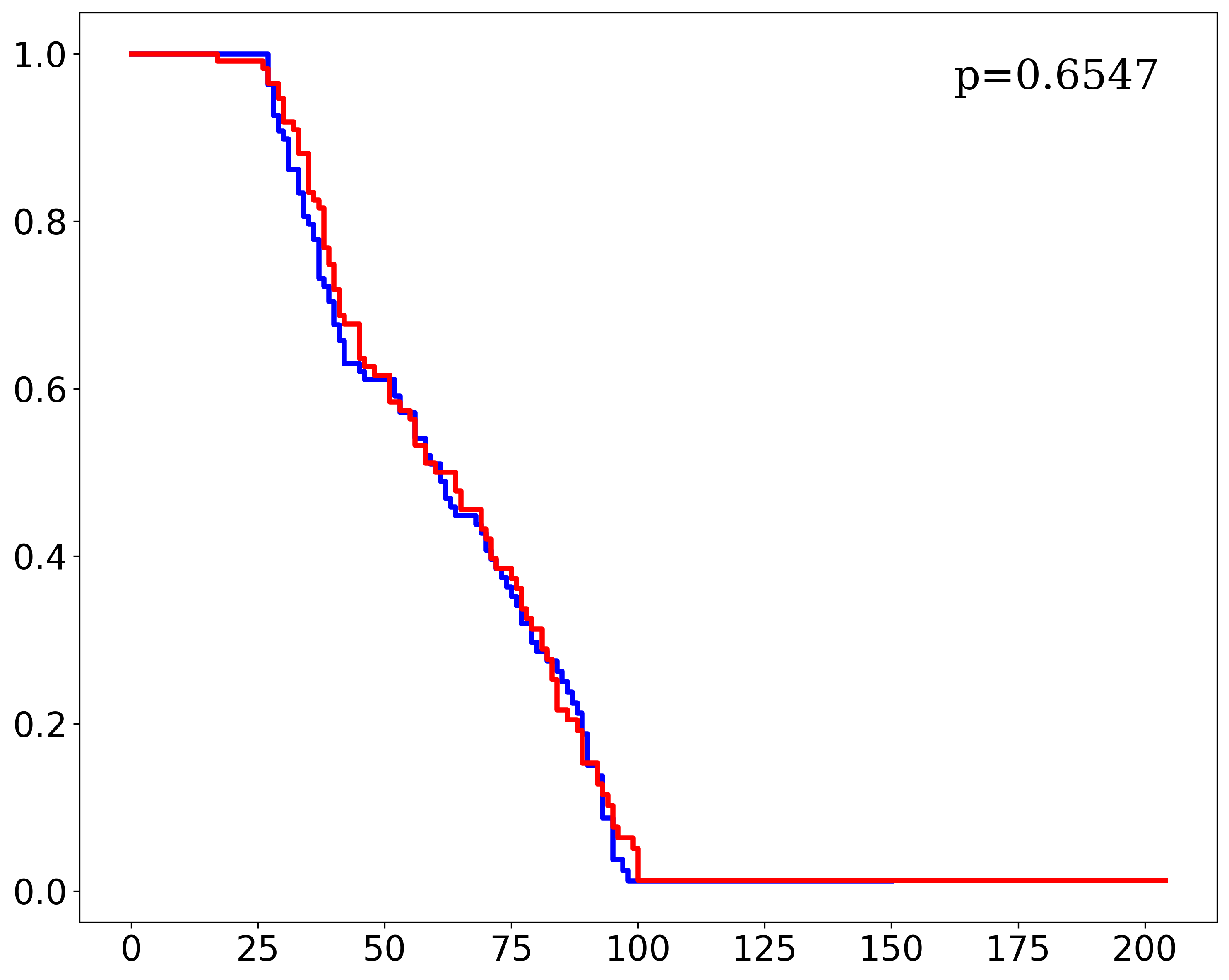}\hfill
        \includegraphics[width=0.19\linewidth]{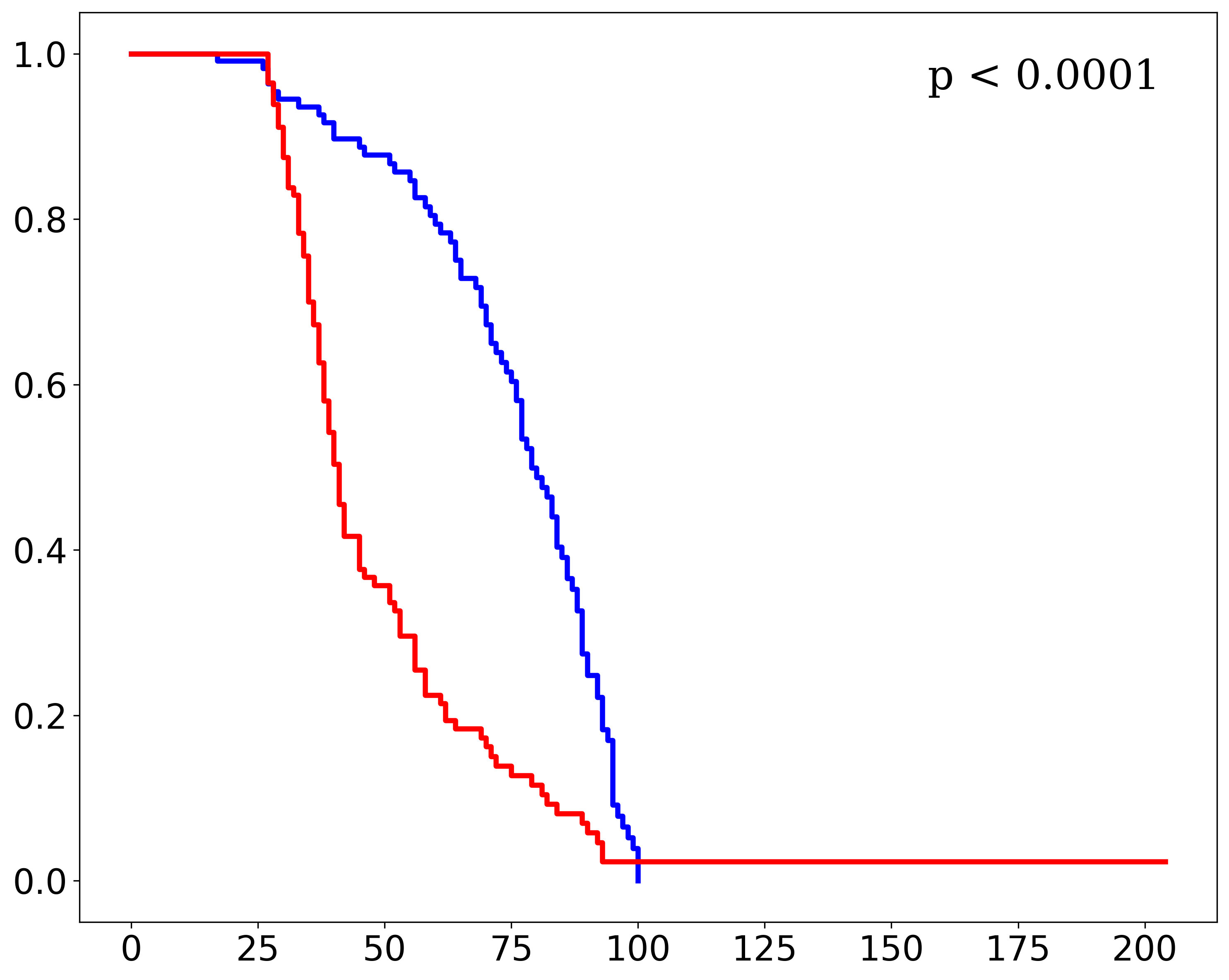}\hfill
        \includegraphics[width=0.19\linewidth]{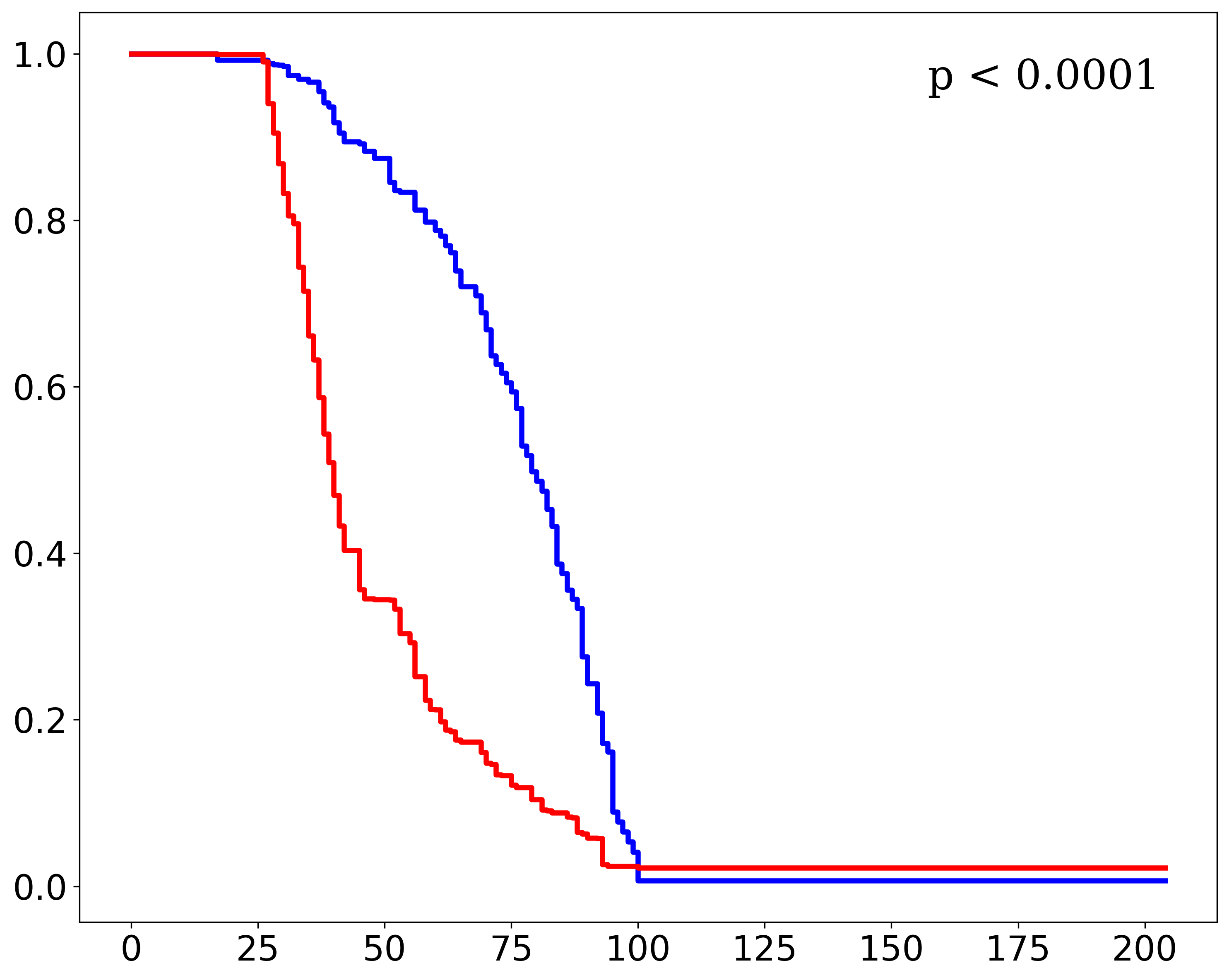}\hfill
        \includegraphics[width=0.19\linewidth]{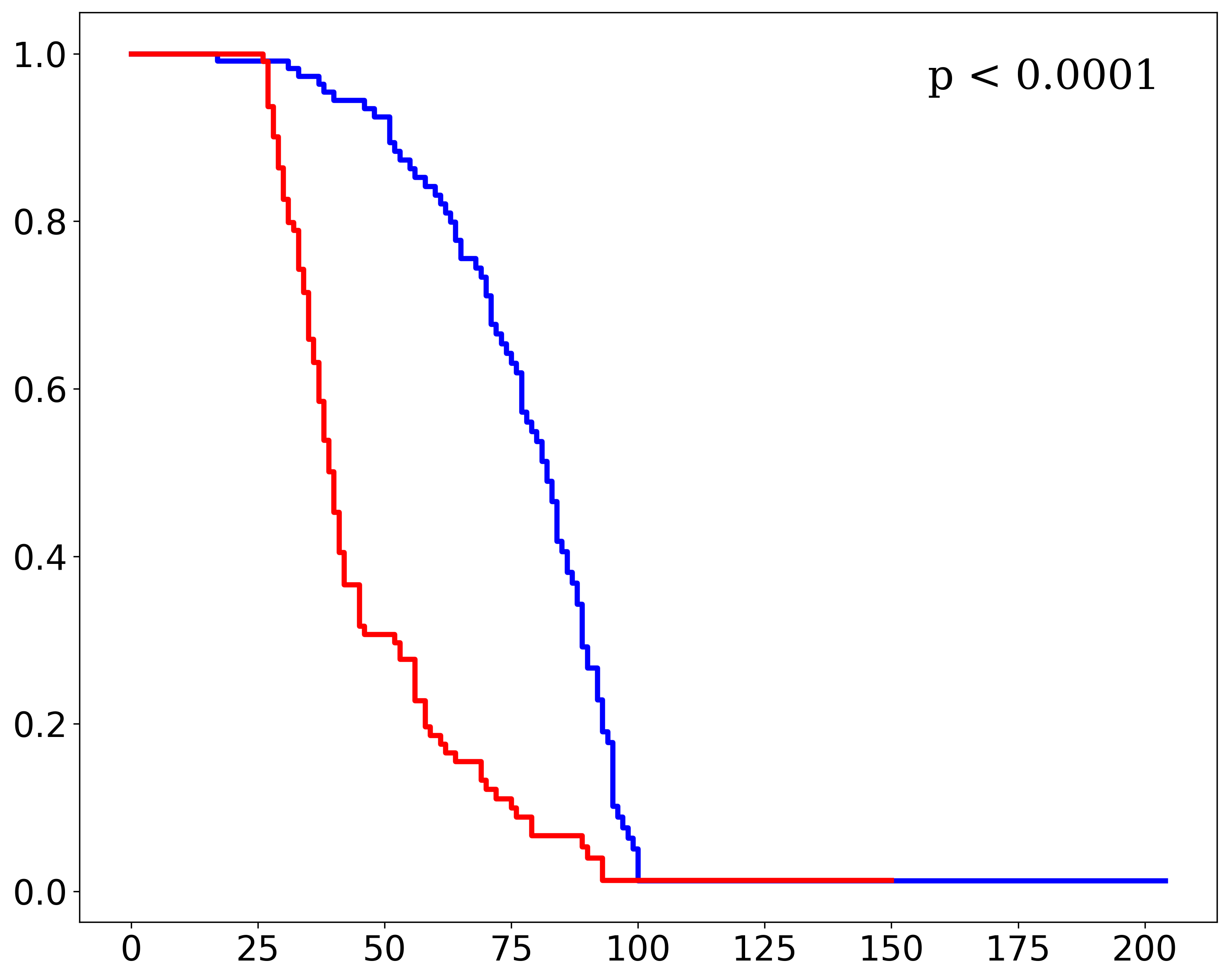}\hfill
        \includegraphics[width=0.19\linewidth]{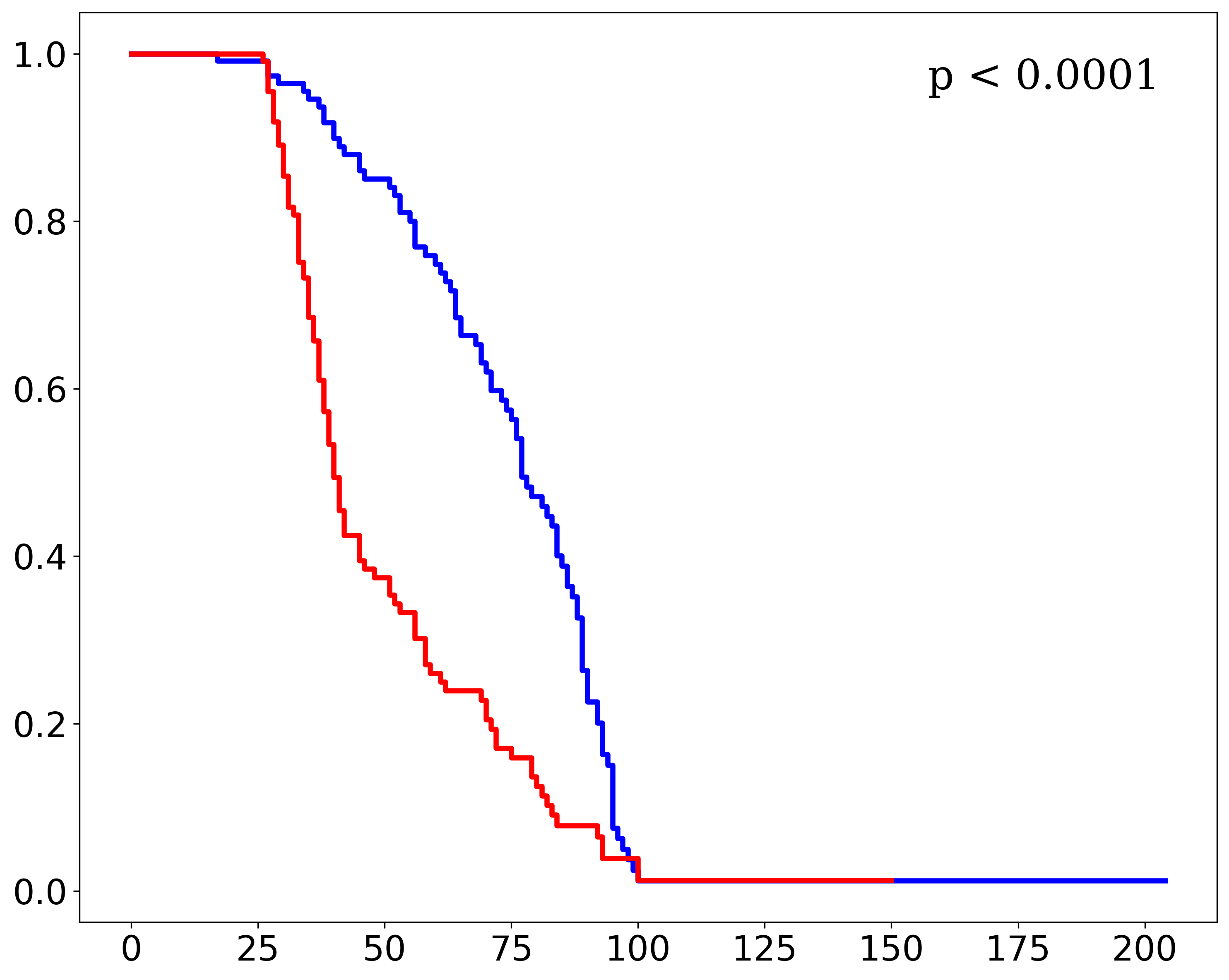}
    \end{minipage}
    
    \vspace{1mm} % Small vertical gap

    % --- ROW 2: OS Curves ---
    % Vertical Label (OS)
    \begin{minipage}{0.03\textwidth}
        \centering
        \rotatebox{90}{OS}
    \end{minipage}
    \hfill
    % Images
    \begin{minipage}{0.96\textwidth}
        \includegraphics[width=0.19\linewidth]{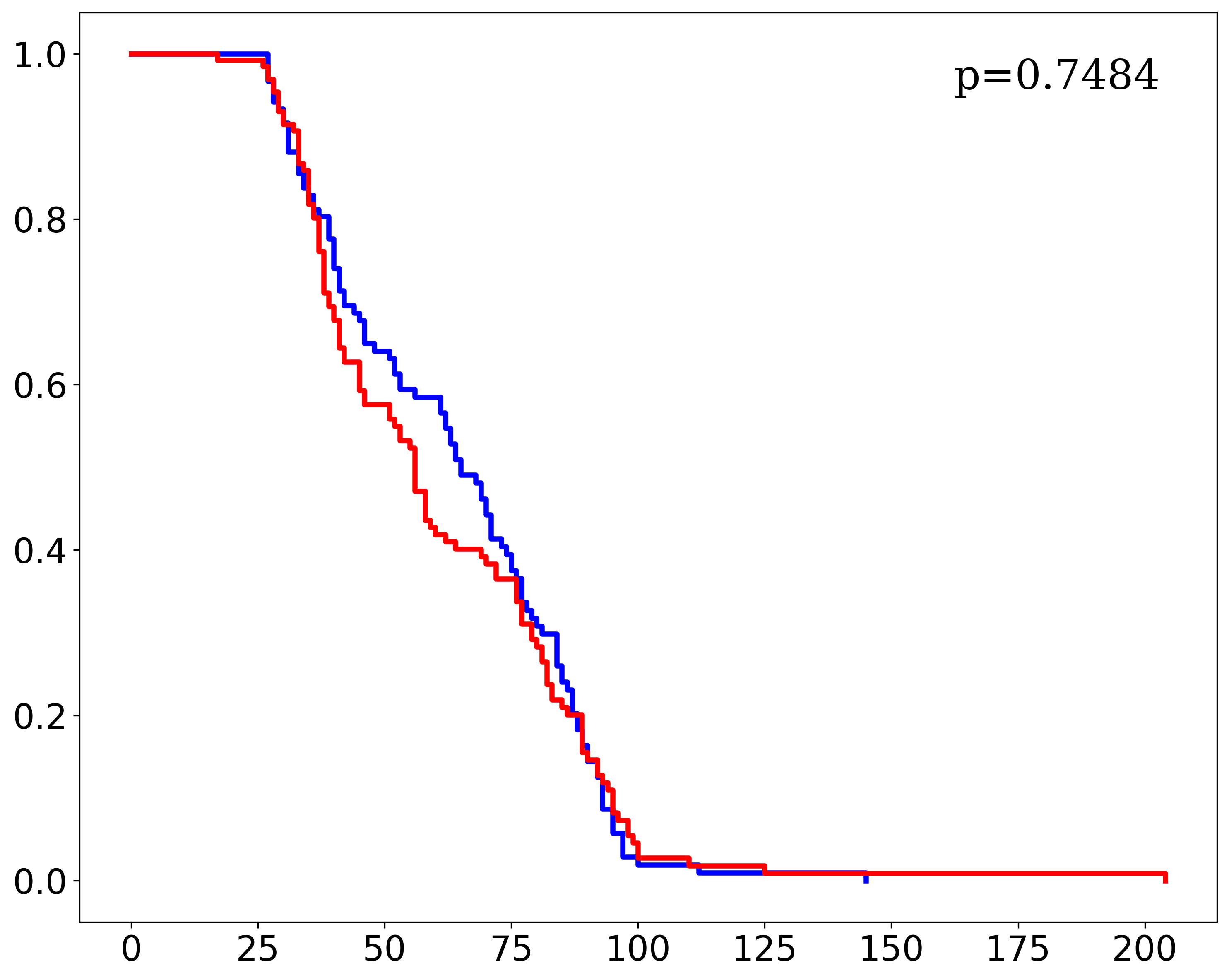}\hfill
        \includegraphics[width=0.19\linewidth]{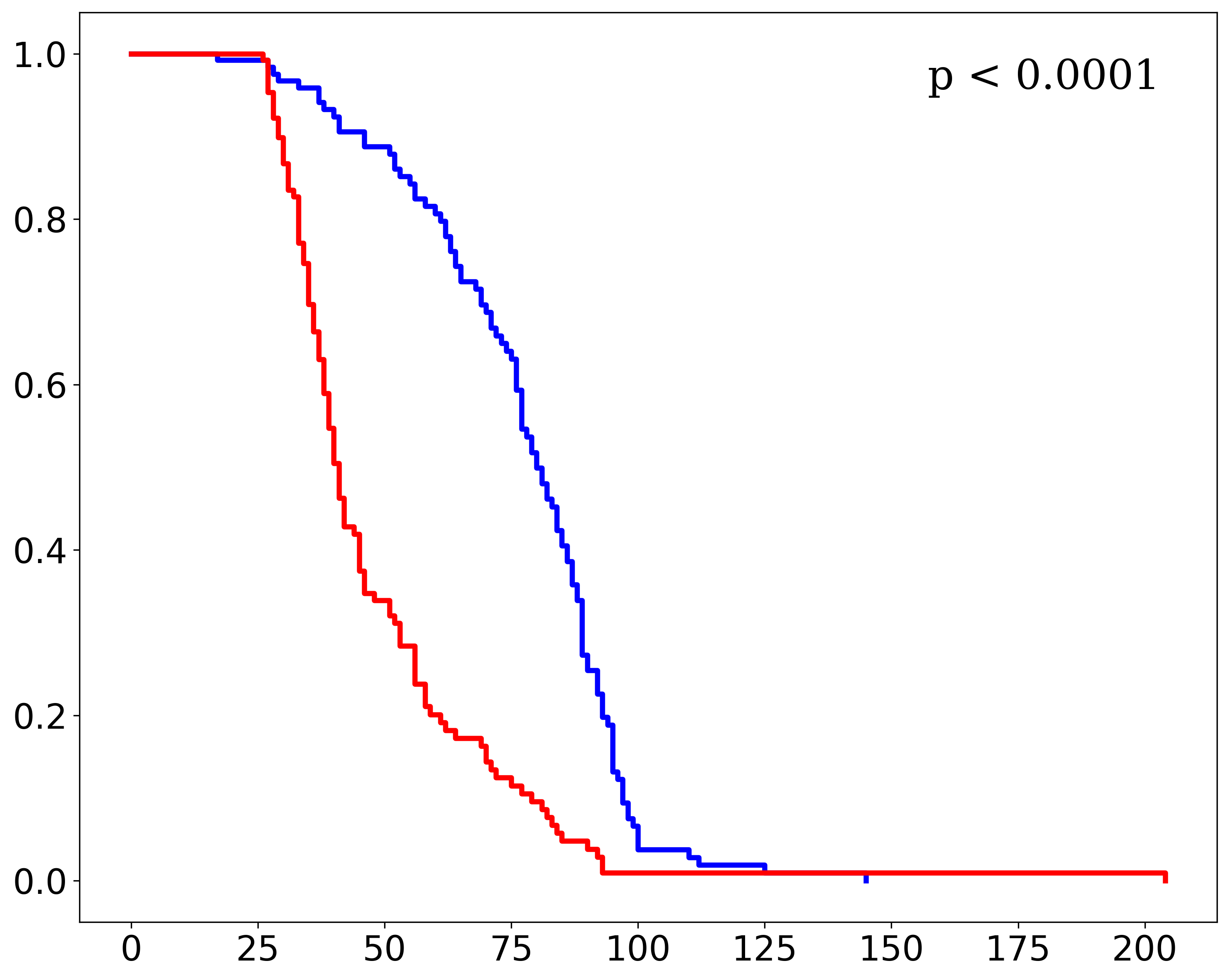}\hfill
        \includegraphics[width=0.19\linewidth]{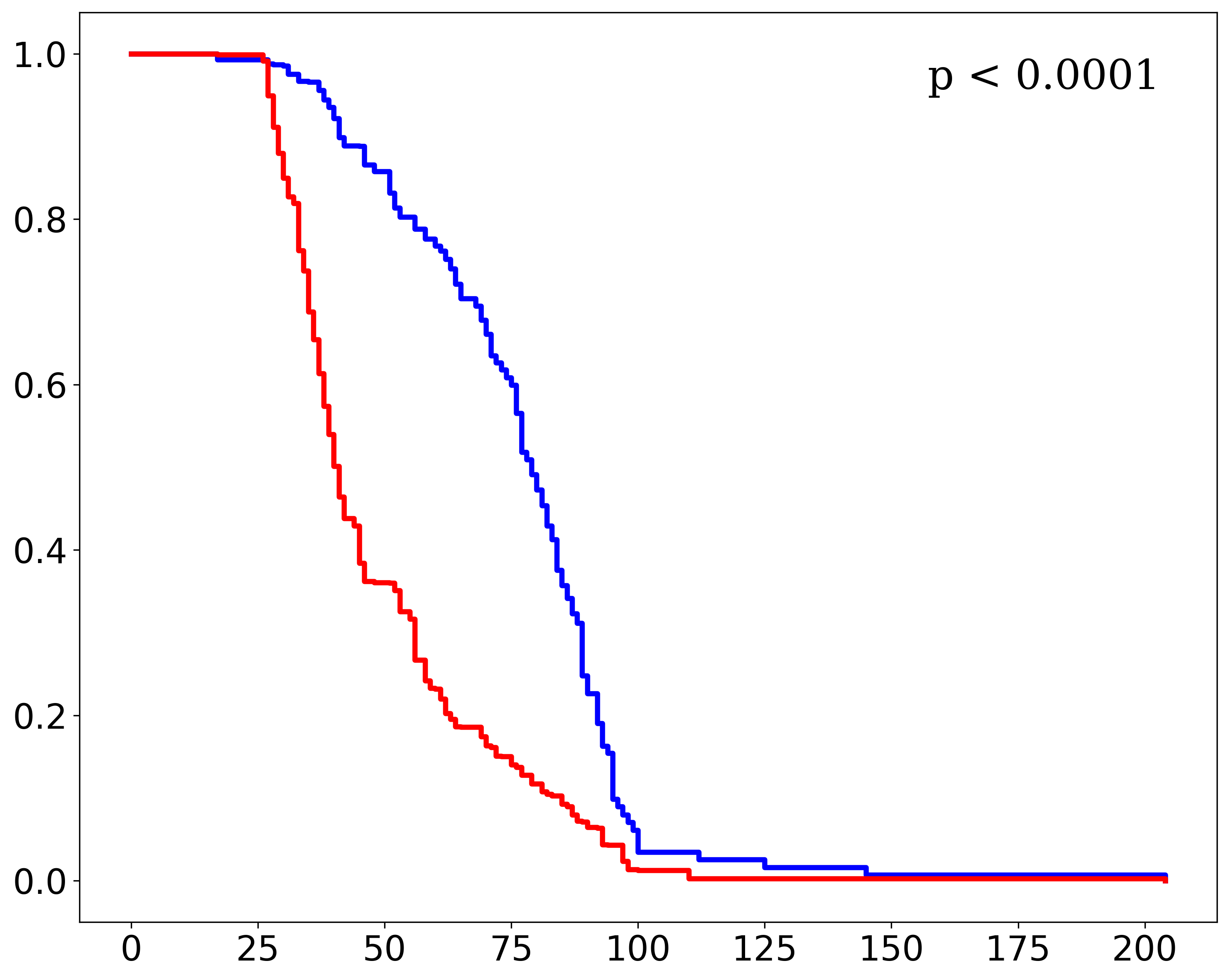}\hfill
        \includegraphics[width=0.19\linewidth]{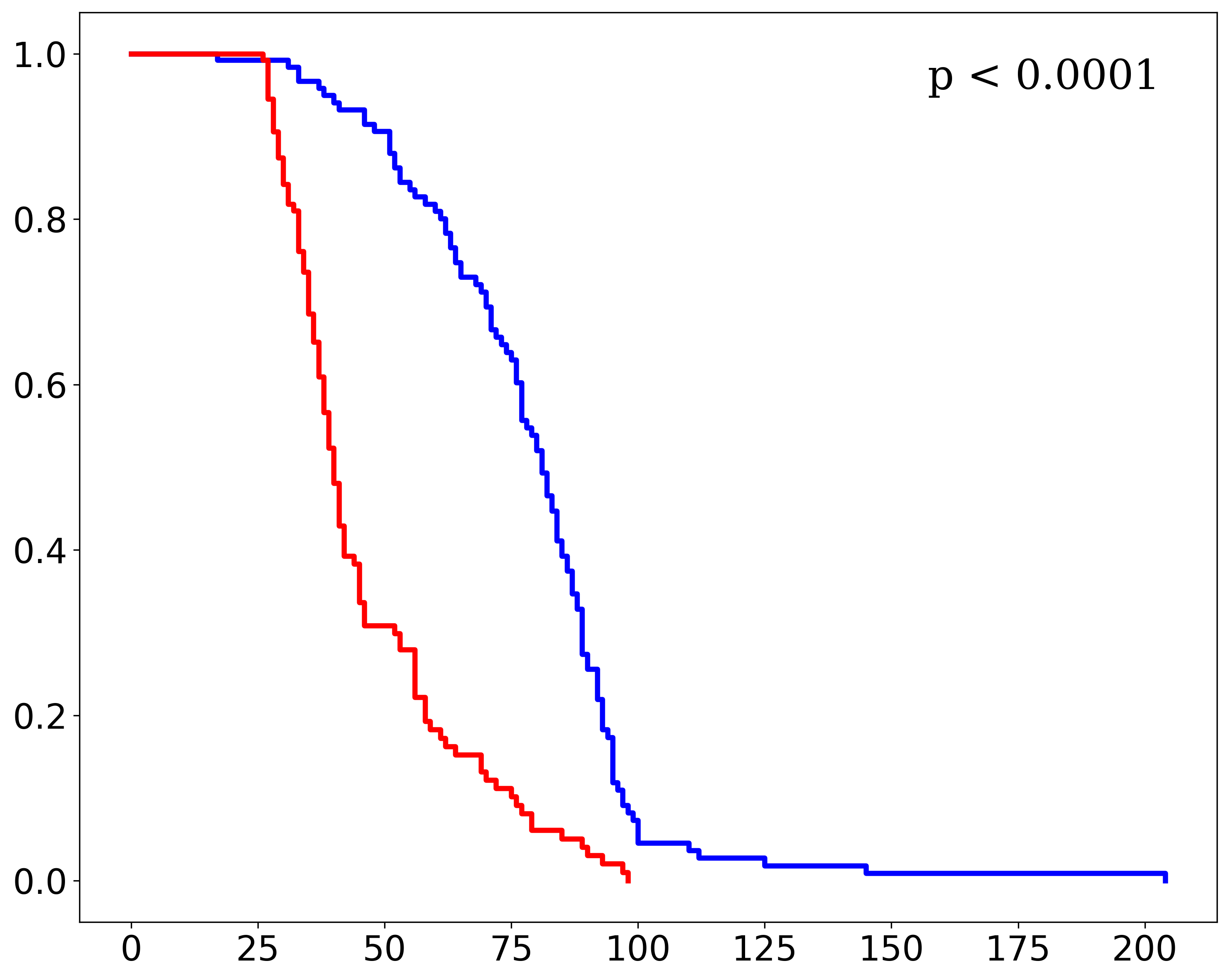}\hfill
        \includegraphics[width=0.19\linewidth]{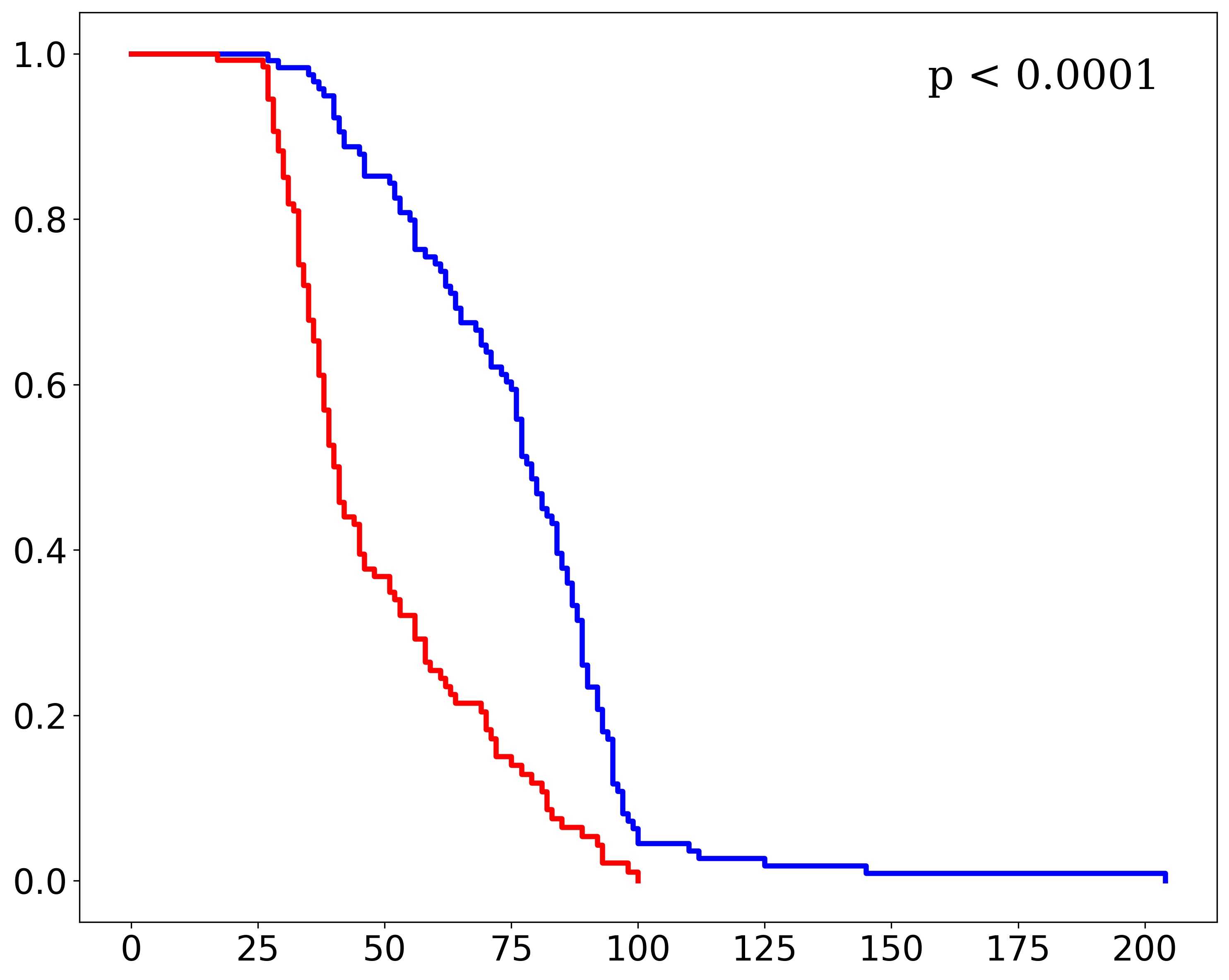}
    \end{minipage}

    \caption{The KM curves of all the compared methods on prognosis metrics (PFS/OS). Predicted risk scores are split into high risk (red) and low risk (blue) based on the median grade.}
    \label{fig:km_curves}
\end{figure*}

\begin{table*}[ht]
\centering
\caption{Comparison with State-of-the-Art Methods. All methods utilize only MRI and Clinical data during inference.}
\label{tab:main_results}

\setlength{\tabcolsep}{16pt}
\renewcommand{\arraystretch}{1.2}

\begin{tabular}{l|cc|cc}
    \toprule
    \multirow{2}{*}{\textbf{Method}} & \multicolumn{2}{c|}{\textbf{Diagnosis (Accuracy)}} & \multicolumn{2}{c}{\textbf{Prognosis (C-Index)}} \\
     & \textbf{Mol. Group} & \textbf{WHO Grade} & \textbf{PFS} & \textbf{OS} \\ \midrule
    
    DeepSurv (Clinical Only) & $0.8587_{\pm0.0489}$ & $0.7042_{\pm0.0480}$ & $0.4739_{\pm0.0414}$ & $0.4956_{\pm0.0337}$ \\
    
    Deep-Fusion & $0.8494_{\pm0.0437}$ & $0.7206_{\pm0.0442}$ & $0.7251_{\pm0.0661}$ & $0.7174_{\pm0.0614}$ \\
    
    AttentionDeepMIL & $0.8108_{\pm0.0545}$ & $0.6986_{\pm0.0837}$ & $\mathbf{0.7671_{\pm0.0292}}$ & $0.7394_{\pm0.0295}$ \\
    
    MCAT & $0.7850_{\pm0.0362}$ & $0.6789_{\pm0.0318}$ & $0.7481_{\pm0.0497}$ & $0.7275_{\pm0.0519}$ \\
    
    \textbf{HyperPriv-EPN (Ours)} & $\mathbf{0.8591_{\pm0.0500}}$ & $\mathbf{0.7208_{\pm0.0656}}$ & $0.7634_{\pm0.0495}$ & $\mathbf{0.7585_{\pm0.0423}}$ \\ \bottomrule
    \end{tabular}%
\end{table*}

\subsection{Datasets and Implementation Details}
\textbf{Clinical Cohort and Data Acquisition.} 
We conducted a comprehensive retrospective study on a multi-center cohort of 311 patients with histopathologically confirmed Posterior Fossa Ependymoma (PF-EPN), treated between May 2000 and November 2022. This represents one of the largest multi-modal datasets for this rare pathology. The inclusion criteria required complete pre-operative multi-modal MRI sequences (T1, T1C, T2, FLAIR, ADC) and structured surgical reports documenting tumor location, texture, and vascularity. For survival analysis (PFS, OS), we utilized a high-quality subset with complete long-term follow-up data.

\textbf{Preprocessing Pipeline.}
Magnetic Resonance Imaging (MRI) data underwent a standardized preprocessing pipeline, including skull stripping, N4 bias field correction, and rigid co-registration to the T1C space using SimpleITK\cite{beare2018simpleitk}. Tumor regions of interest (ROIs) were automatically segmented using a pre-trained nnU-Net\cite{isensee2021nnu} and verified by senior neuroradiologists. 
Unlike standard approaches, we extracted visual features using a 3D CNN backbone enhanced with Hierarchical Self-Supervised Learning (HSSL) to capture lesion-specific representations via InfoNCE contrastive pre-training. 
For textual data, we employed the Qwen Large Language Model (LLM) to extract dense semantic embeddings and sparse clinical concepts, capturing critical phenotypic descriptions (e.g., "necrosis") often missed by imaging alone.

\subsection{Experimental Setup and Evaluation Metrics}

\textbf{Clinical Tasks.} We evaluate performance on four standard neuro-oncology tasks:
\begin{itemize}
    \item Molecular Grouping (Classification): Predicting the biological subtype (Posterior Fossa A vs. B), which is crucial as PFA tumors typically carry a poorer prognosis than PFB.
    \item WHO Grading (Classification): Assigning the tumor grade (Grade II vs. III) according to the 2021 WHO CNS classification standards.
    \item Overall Survival (OS): Regression task predicting the time (in months) from surgery to patient death.
    \item Progression-Free Survival (PFS): Regression task predicting the time from surgery to tumor recurrence or progression.
\end{itemize}

\textbf{Evaluation Metrics.}
\begin{itemize}
    \item Accuracy (ACC): For diagnostic classification (Group/WHO), we report standard accuracy.
    \item Concordance Index (C-Index)\cite{harrell1996multivariable}: For survival analysis, we utilize the C-Index, which generalizes the Area Under the ROC Curve (AUC) to censored data. It measures the probability that, for a random pair of patients, the model correctly predicts which patient will survive longer. $c = 0.5$ implies random guessing, while $c = 1.0$ implies perfect ranking.
    \item Kaplan-Meier (KM) Analysis \cite{kaplan1958nonparametric}: To assess clinical utility, we stratify patients into low-risk and high-risk groups based on the median predicted risk score. We then visualize the survival probability over time using KM curves and quantify the statistical significance of the separation using the Log-Rank test ($p$-value).\cite{mantel1966evaluation}
\end{itemize}

\subsection{Main Results}
Table \ref{tab:main_results} summarizes the performance comparison on the testing cohort. The results reveal distinct trade-offs inherent in existing methods, which HyperPriv-EPN successfully resolves.

\textbf{The Clinical Baseline Trade-off.}
DeepSurv demonstrates surprisingly strong diagnostic accuracy (Group: 0.8587, WHO: 0.7042), suggesting that clinical variables like age and tumor invasion are highly predictive of the diagnosis. However, it fails catastrophically on prognosis (PFS: 0.4739, OS: 0.4956), performing no better than random guessing. This confirms that while clinical metadata is sufficient for classification, it lacks the granularity to predict long-term survival outcomes.

\textbf{The SOTA Prognosis Trade-off.}
Advanced deep learning baselines (AttentionDeepMIL, MCAT) reverse this trend. While they achieve superior prognostic performance (AttentionDeepMIL PFS: 0.7671), they paradoxically suffer a degradation in diagnostic accuracy compared to the simple clinical baseline (e.g., MCAT WHO: 0.6789 vs. DeepSurv WHO: 0.7042). This suggests that complex deep learning models may overfit to visual signals at the expense of robust clinical priors.

\textbf{HyperPriv-EPN: The "All-Rounder".}
Our proposed framework uniquely bridges this gap. By utilizing Student-Teacher distillation, HyperPriv-EPN retains the high diagnostic accuracy of clinical models (Group: 0.8591, WHO: 0.7208) while simultaneously achieving state-of-the-art prognostic stratification. Notably, we outperform AttentionDeepMIL on OS (0.7585 vs. 0.7394) and achieve statistically comparable performance on PFS (0.7634). This demonstrates that our method effectively "hallucinates" the missing semantic context required for holistic patient assessment, delivering optimal performance across both diagnosis and prognosis tasks.

\textbf{Visual Stratification Validation.}
We further validate these findings via Kaplan-Meier analysis (Fig. \ref{fig:km_curves}). Consistent with the quantitative metrics, the clinical-only DeepSurv baseline fails to stratify patients ($p > 0.05$). In contrast, all deep learning approaches, including MCAT and AttentionDeepMIL, achieve statistically significant separation ($p < 0.0001$). Crucially, HyperPriv-EPN maintains this robust stratification capability, producing clean separation curves comparable to the multimodal MCAT baseline. This visually confirms that our distillation strategy successfully preserves the prognostic discriminability of the privileged teacher, allowing for distinct risk stratification even without access to surgical reports at inference time.

\begin{table}[h]
    \centering
    \caption{Ablation study of HyperPriv-EPN components.}
    \label{tab:ablation}
    \resizebox{\columnwidth}{!}{%
    \begin{tabular}{l|cc|cc}
    \toprule
    \multirow{2}{*}{\textbf{Method}} & \multicolumn{2}{c|}{\textbf{Diagnosis (Accuracy)}} & \multicolumn{2}{c}{\textbf{Prognosis (C-Index)}} \\
     & \textbf{Group} & \textbf{WHO} & \textbf{PFS} & \textbf{OS} \\ \midrule
    
    w/o Hypergraph & 0.8587 & 0.7126 & 0.7153 & 0.7071 \\
    
    w/o Knowledge Distillation & 0.8494 & 0.7206 & 0.7251 & 0.7174 \\
    
    w/o SSL (Raw MRI signals) & 0.7463 & 0.6301 & 0.6716 & 0.6617 \\ 
    
    \textbf{HyperPrivEPN (Ours)} & \textbf{0.8591} & \textbf{0.7208} & \textbf{0.7634} & \textbf{0.7585} \\ \midrule
    
    HyperPrivEPN (Teacher) & 0.8687 & 0.7300 & 0.7823 & 0.7644 \\ 
    
    \bottomrule
    \end{tabular}%
    }
\end{table}

\subsection{Ablation Study}
We further validated our architectural contributions through an ablation study (Table \ref{tab:ablation}).

\textbf{Impact of Hypergraph and Distillation.}
Removing the hypergraph structure or the distillation mechanism (i.e., the "Deep Fusion" variant) leads to a significant drop in prognostic performance ($\sim$5\% decrease in PFS/OS), though diagnostic accuracy remains stable. This confirms that the severed graph strategy is specifically responsible for transferring the complex survival-related semantic knowledge from the teacher.

\textbf{Importance of SSL Pre-training.}
The most critical component proved to be the Hierarchical Self-Supervised Learning (HSSL) encoder. Replacing our InfoNCE-pretrained embeddings with standard ResNet features ("w/o SSL") caused a drastic $\sim$10\% drop across all metrics. This indicates that without robust, contrastive node embeddings, the hypergraph cannot effectively model inter-patient similarities.

\textbf{Teacher vs. Student Gap.}
Finally, comparing the Student against the privileged Teacher (which accesses text during inference) reveals a negligible performance gap ($<$1\% in Diagnosis, $\sim$1-2\% in Prognosis). This marginal difference validates the efficacy of our distillation: the Student has successfully internalized the vast majority of the Teacher's privileged insights, rendering the text reports largely redundant at inference time.

\section{Conclusion}
In this paper, we introduced HyperPriv-EPN, a novel hypergraph distillation framework designed to address the inference-time modality gap in Ependymoma prognosis. By integrating a Severed Graph Strategy with LUPI, our method effectively distills high-order semantic insights from surgical reports into an MRI-only student model. Experimental results on a multi-center cohort demonstrate that HyperPriv-EPN significantly outperforms state-of-the-art multimodal baselines in diagnostic accuracy while achieving robust risk stratification comparable to models with full data access. Future efforts will focus on extending this hypergraph distillation paradigm to other rare pathologies beyond Ependymoma and validating the framework on larger, external multi-institutional cohorts.

\end{document}